\documentclass{article}

\usepackage{arxiv}

\usepackage[utf8]{inputenc} 
\usepackage[T1]{fontenc}    
\usepackage{hyperref}       
\usepackage{url}            
\usepackage{booktabs}       
\usepackage{amsfonts}       
\usepackage{microtype}      
\usepackage{graphicx}
\graphicspath{ {./images/} }
\usepackage{amsmath}
\usepackage{multirow}
\usepackage{textcomp}
\usepackage[numbers,sort]{natbib}
\usepackage[separate-uncertainty=true]{siunitx}
\usepackage[capitalize]{cleveref}
\usepackage{xspace}
\usepackage{bm}
\usepackage{array}

\newcommand{\muxsf}{\ensuremath{\widetilde{\mu}_{x\!_s}}\xspace}
\newcommand{\sigxsf}{\ensuremath{\widetilde{\sigma}_{x\!_s}}\xspace}
\newcommand{\mulf}{\ensuremath{\widetilde{\mu}_{\scriptscriptstyle{l}}}\xspace}
\newcommand{\siglf}{\ensuremath{\widetilde{\sigma}_{\scriptscriptstyle{l}}}\xspace}
\newcommand{\csl}{\ensuremath{c_{\scriptscriptstyle{s}\!,\scriptscriptstyle{l}}}\xspace}
\newcommand{\muxtf}{\ensuremath{\widetilde{\mu}_{x\!_t}}\xspace}
\newcommand{\sigxtf}{\ensuremath{\widetilde{\sigma}_{x\!_t}}\xspace}
\newcommand{\Ullf}{\ensuremath{\bm{\widetilde{U}_{\scriptscriptstyle{ll}}}}\xspace}
\newcommand{\Unf}{\ensuremath{\bm{\widetilde{U}_{\scriptscriptstyle{n}}}}\xspace}
\newcommand{\Epotf}{\ensuremath{\widetilde{E}_{\scriptscriptstyle{\text{potts}}}}\xspace}
\newcommand{\Ebanf}{\ensuremath{\widetilde{E}_{\scriptscriptstyle{\text{ban}}}}\xspace}
\newcolumntype{C}[1]{>{\centering\arraybackslash}m{#1}}

\title{Unsupervised Segmentation of Micro-CT Scans of Polyurethane Structures By Combining Hidden-Markov-Random Fields and a U-Net}

\renewcommand{\shorttitle}{Unsupervised Segmentation of Micro-CT Scans Using an HMRF-UNet}

\usepackage{authblk}

\newbox{\orcid}\sbox{\orcid}{\includegraphics[scale=0.06]{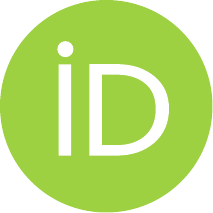}}
\author[1,2,3,4]{%
	\href{https://orcid.org/0009-0003-2436-5928}{\usebox{\orcid}\hspace{1mm}Julian~Grolig\thanks{\texttt{julian.grolig@kit.edu}}}%
}
\author[1,2]{%
	\href{https://orcid.org/0000-0002-8093-6356}{\usebox{\orcid}\hspace{1mm}Lars~Griem}%
}
\author[1,5]{%
	\href{https://orcid.org/0000-0002-9756-646X}{\usebox{\orcid}\hspace{1mm}Michael~Selzer}%
}
\author[3,4]{%
	\href{https://orcid.org/0000-0002-6730-9462}{\usebox{\orcid}\hspace{1mm}Hans-Ulrich~Kauczor}%
}
\author[3,4]{%
	\href{https://orcid.org/0000-0003-2068-1184}{\usebox{\orcid}~\hspace{1mm}Simon~M.~F.~Triphan}%
}
\author[1,2,5]{%
	\href{https://orcid.org/0000-0002-3768-3277}{\usebox{\orcid}\hspace{1mm}Britta~Nestler}%
}
\author[1,2]{%
	\href{https://orcid.org/0000-0002-4833-1306}{\usebox{\orcid}\hspace{1mm}Arnd~Koeppe\thanks{\texttt{arnd.koeppe@kit.edu}}}%
}

\affil[1]{Institute of Nanotechnology (INT), Karlsruhe Institute of Technology (KIT), Eggenstein-Leopoldshafen, Germany}
\affil[2]{Institute for Applied Materials - Microstructure Modelling and Simulation (IAM-MMS), Karlsruhe Institute of Technology (KIT), Karlsruhe, Germany}
\affil[3]{Diagnostic and Interventional Radiology, University Hospital of Heidelberg, Heidelberg, Germany}
\affil[4]{Translational Lung Research Center Heidelberg (TLRC), German Center for Lung Research (DZL), Heidelberg, Germany}
\affil[5]{Institute of Digital Materials Science (IDM), Karlsruhe University of Applied Sciences, Karlsruhe, Germany}

\begin{document}
\maketitle

\begin{abstract}

Extracting digital material representations from images is a necessary prerequisite for a quantitative analysis of material properties. Different segmentation approaches have been extensively studied in the past to achieve this task, but were often lacking accuracy or speed. With the advent of machine learning, supervised convolutional neural networks (CNNs) have achieved state-of-the-art performance for different segmentation tasks. However, these models are often trained in a supervised manner, which requires large labeled datasets. Unsupervised approaches do not require ground-truth data for learning, but suffer from long segmentation times and often worse segmentation accuracy. Hidden Markov Random Fields (HMRF) are an unsupervised segmentation approach that incorporates concepts of neighborhood and class distributions. We present a method that integrates HMRF theory and CNN segmentation, leveraging the advantages of both areas: unsupervised learning and fast segmentation times. We investigate the contribution of different neighborhood terms and components for the unsupervised HMRF loss. We demonstrate that the HMRF-UNet enables high segmentation accuracy without ground truth on a Micro-Computed Tomography (\textmu CT) image dataset of Polyurethane (PU) foam structures. Finally, we propose and demonstrate a pre-training strategy that considerably reduces the required amount of ground-truth data when training a segmentation model.
\end{abstract}


\section{Introduction}
\label{sec:introduction}

Micro-Computed Tomography (\textmu CT) is a commonly used imaging technique to analyze the microstructure of materials. Materials often consist of several components, which are often only discernible on the microscale. The components of a foam structure are, for example, the solid matrix phase and the air phase. Before any quantitative analysis of a material can be conducted, the different components have to be distinguished. A common approach to identify and cluster unique components in images is called segmentation.
Segmentation can be defined as the task of grouping voxels in an image into groups by meaning. Segmentation approaches can be classified into four main categories: threshold-based, region-based, model-based, and pixel-classification techniques. \cite{gordillo_state_2013}

In threshold-based methods, one or more thresholds for pixel values are determined to divide an image into different classes \cite{otsu_threshold_1979,ridler_picture_1978}. Region-based approaches, such as region growing, start from seed-points before connecting and grouping voxels, if they have similar properties and are spatially linked \cite{adams_seeded_1994,hojjatoleslami_region_1998}. Other approaches combine threshold- and region-based methods to segment structures in images \cite{bogunia_microstructure_2022}. Model-based approaches usually combine (statistical) shape models and registration approaches to fit image structures to a template shape \cite{neumann_statistical_1998,kelemen_elastic_1999}. Another recent model-based approach used a combination of skeletonization and pruning for microstructure segmentation \cite{deshpandeSkeletonizationBasedImage2025}. The last category, pixel-classification techniques, combines both unsupervised methods, such as Self-Organizing-Maps (SOM) \cite{kohonen_self-organizing_1990} or Hidden Markov Random Fields (HMRF) \cite{zhang_segmentation_2001,panicGuideUnsupervisedImage2024}, weakly supervised methods \cite{na_unified_2023}, and supervised methods such as convolutional neural networks (CNNs) \cite{ronneberger_u-net_2015,milletari_v-net_2016}.
Supervised CNNs, e.g., the commonly used U-Net model, generate state-of-the-art segmentations \cite{furat_machine_2019,isensee_nnu-net_2021,astley_implementable_2023,medghalchiAutomatedSegmentationLarge2024}. However, supervised models require a lot of annotated ground-truth data, which is often not available, because it is unknown or expensive to collect. Another drawback of using ground-truth data is the potential for added error if the data was incorrectly collected. Hidden Markov Random Fields are a widely used method for unsupervised segmentation, and are often iteratively optimized using the time-consuming combination of the expectation-maximization (EM) and iterated conditional modes (ICM) algorithms \cite{zhang_segmentation_2001,gordillo_state_2013}. Recently, evolutionary algorithms (EAs) were shown to be good alternatives for solving the NP-Hard HMRF optimization problem, which is non-linear, complex, and has several local minima~\cite{guerrout_hidden_2020,mahiou_taguchi_2023}. However, such evolutionary algorithms must be rerun for every new image to be segmented. The time required for segmentation using EAs is comparable to that of the EM-ICM techniques, but both are still significantly slower than predictions by trained CNNs. Furthermore, HMRF evolutionary methods require excessive computational resources for each new segmentation. The final drawback of HMRF segmentation is that, due to the one-shot learning, no experience from previous segmentations can be learned. 

To overcome these drawbacks, we propose the novel unsupervised segmentation network HMRF-UNet, which combines the HMRF concept with the common U-Net structure for the unsupervised segmentation of images. The HMRF-UNet uses HMRF-based losses to train the U-Net network and thereby combines the advantages of both methods. The HMRF-UNet learns without any ground-truth, but also learns from known segmentations, and generates fast and resource-saving segmentation predictions. We demonstrate the performance and influence of different neighborhood terms and weights on segmentation results for a dataset of Polyurethane foam \textmu CT images and compare them to the segmentation results of a supervised U-Net.

\section{Methods}
\label{sec:methods}

\subsection{HMRF segmentation theory}

Let the image $\bm{y}$ be a realization of a random field $\bm{Y}$, representing the observed pixel intensities. Let the segmentation $\bm{x}$ be a configuration of a second random field $\bm{X}$, which assigns a label (e.g., class or region) to each voxel in the image. Then we can call $\bm{X}$ a Hidden-Markov-Random-Field of $\bm{Y}$.
Let $s$ be a voxel on the image grid $S$ and $N_s$ be the neighborhood clique of $s$ from the neighborhood system $N$, which connects all voxels in $S$. Then $y_s$ can be defined as the value (e.g. density in a \textmu CT image) of the voxel $s$ within the image $\bm{y}$. The assigned class of voxel $s$ can be described by $x_s \in L$, where $L$ is the set of possible classes in the image. The image can be represented by a Gaussian mixture model, where each class $l\in L$ in the image $\bm{y}$ is represented by a Gaussian distribution with the parameter set $\theta_l=\{\mu_l,\sigma_l\}$ with mean $\mu_l$ and standard deviation $\sigma_l$. The posterior distribution is defined as:

\begin{equation}
    P(y_s|\theta_l)=\frac{1}{\sqrt{2\pi\sigma_l^2}}\exp\!\left[-\frac{(y_s-\mu_l)^2}{2\sigma_l^2}\right]
\end{equation}
and the conditional likelihood energy function $\bm{U_{ll}}(\bm{y}|\bm{x})$ is given by 
\begin{equation}
    \bm{U_{ll}}(\bm{y}|\bm{x})=\sum_{s \in S}\left(\frac{(y_s-\mu_{x_s})^2}{2\sigma_{x_s}^2}+\ln{\sigma_{x_s}}\right).
    \label{eq:cond_llhood}
\end{equation}

Based on a possible segmentation $\bm{x}$, a stabilizing a priori probability term can be defined in the form of an energy function $\bm{U_n}(\bm{X}=\bm{x})$, which takes into account all neighborhood cliques $N_s$ in image $\bm{Y}$  \cite{zhang_segmentation_2001}:
\begin{equation}
    \bm{U_n}(\bm{X}=\bm{x})=\sum_{s\in S}E(x_s)
    \label{eq:neigh_energy}
\end{equation}
where $E(x_s)$ is the neighborhood penalty of the neighborhood $N_s$. One common neighborhood penalty is the Potts neighborhood~\cite{ait-aoudia_evaluation_2011,zhang_segmentation_2001}, which only takes into account the class of neighboring voxels:
\begin{equation}
    E_{\text{potts}}(x_s) = \alpha\sum_{t\in N_s} \left(1-2\delta_{x_sx_t}\right).
    \label{eq:potts_penalty}
\end{equation}
Here, $\delta_{x_sx_t}$ is the Kronecker delta between $x_s$ and $x_t$, and $\alpha$ is a weight for the neighborhood term. Another possible neighborhood penalty term, which also includes Gaussian distribution properties of the neighbors' classes, was defined by \citet{banerjee_spatially_2020}:
\begin{equation}
    E_{\text{ban}}(x_s) = \frac{\alpha_s}{2|N_s|}\sum_{t\in N_s}\left((\mu_{x_s}-\mu_{x_t})^2\left(\frac{1}{\sigma_{x_s}^2}+\frac{1}{\sigma_{x_t}^2}\right)-1\right)
    \label{eq:banerjee_penalty}
\end{equation}
 where $\alpha_s$ is a voxel-specific weight factor, which is determined by 
\begin{equation}
 	\alpha_s=
 	\begin{cases}
        \begin{aligned}
 		2\alpha, \quad &\text{if} \left(r_{s(1)}^{(2)} - r_{s(2)}^{(2)} \right) > T \quad &\text{and} \quad \left(r_{s(1)}^{(1)} - r_{s(2)}^{(1)} \right) > T \\
 		\alpha, \quad &\text{if} \left(r_{s(1)}^{(2)} - r_{s(2)}^{(2)} \right) > T \quad &\text{and} \quad \left(r_{s(1)}^{(1)} - r_{s(2)}^{(1)} \right) \leq T \\
 		\alpha, \quad &\text{if} \left(r_{s(1)}^{(2)} - r_{s(2)}^{(2)} \right) \leq T \quad &\text{and} \quad \left(r_{s(1)}^{(1)} - r_{s(2)}^{(1)} \right) > T \\
 		0, \quad &\text{otherwise} &\\
        \end{aligned}
 	\end{cases}
    \label{eq:custom_neigh_weight}
 \end{equation}
where $T$ is a threshold and $r_{s_i}^{(k)}$ is the relative number of neighbors belonging to the $i^{th}$ most common class for the $k$-order neighborhood. A detailed explanation can be found in ~\citet{banerjee_spatially_2020}. 

The optimal segmentation $\bm{\hat{x}}$ can be found by minimizing the sum of both energy terms in \cref{eq:cond_llhood,eq:neigh_energy}:
\begin{equation}
    \bm{\hat{x}}=\arg_{min}\left(\sum_{s \in S}\left(\frac{(y-\mu_{x_s})^2}{2\sigma_{x_s}^2}+\ln{\sigma_{x_s}}+E(x_s)\right)\right).
\end{equation}

\subsection{Novel fuzzy HMRF penalty formulations for neural networks}
The HMRF segmentation task is usually described for the discrete case. However, the prediction of a CNN for segmentation is usually a fuzzy confidence map $\mathbf{c}$, which represents the probability that the voxel belongs to a certain class. The word fuzzy in this context means that a voxel is not assigned to a single class, but to multiple classes using a confidence value for each class. Therefore, we propose a reformulation of the discrete energy functions for the fuzzy case, where the confidences weight the contributions to the energy functions. The mean and standard deviation for each class were calculated using the weighted mean approach, where the value of each voxel contributes to the fuzzy mean \mulf and the fuzzy standard deviation \siglf of a class based on its class probability \csl.

\begin{equation}
	\mulf= \frac{\sum\limits_{s \in S} \left( \csl \odot y_s \right)}{|\csl|}
\end{equation}
\begin{equation}
	\siglf=\sqrt{\frac{\sum\limits_{s \in S} \left( \csl \odot (y_s-\mu_l)^2 \right)}{|\csl|}}
\end{equation}
Here, $\odot$ is the Hadamard product and $|\csl|$ is the sum of all confidence values for class $l$.
With the fuzzy mean and standard deviation for each class, we can calculate the fuzzy mean \muxsf and standard deviation \sigxsf belonging to a voxel $s$:
\begin{equation}
    \muxsf=\sum_{l\in L}(\csl\cdot\mulf)
\end{equation}
\begin{equation}
    \sigxsf=\sum_{l\in L}(\csl\cdot\siglf).
\end{equation}
Using this definition, we can reformulate \cref{eq:cond_llhood} as:
\begin{equation}
    \Ullf(\bm{y}|\bm{x})=\sum_{s \in S}\left(\frac{(y_s-\muxsf)^2}{2\sigxsf^2}+\ln{\sigxsf}\right).
    \label{eq:cond_llhood_fuzzy}
\end{equation}
The Potts neighborhood energy term from \cref{eq:potts_penalty} was also reformulated to work for fuzzy labels. Instead of calculating the number of dissimilar labels inside the neighborhood, the average Euclidean distance between the confidence vector $\mathbf{c}_s$ of the voxel $ x_s $ and the confidence vectors $\mathbf{c}_t$ of all its neighbors $ x_t $ was calculated:
\begin{equation}
	\Epotf(x_s)=\frac{\alpha}{|N_s|}\sum_{t\in N_s} \lVert\mathbf{c}_s-\mathbf{c}_t \rVert^2 \;.
    \label{eq:potts_penalty_fuzzy}
\end{equation}
The Banerjee neighborhood energy term from \cref{eq:banerjee_penalty} was also reformulated for the fuzzy case:
\begin{equation}
    \Ebanf(x_s) = \frac{\alpha}{2|N_s|}\sum_{t\in N_s}\left((\muxsf-\muxtf)^2\left(\frac{1}{\sigxsf^2}+\frac{1}{\sigxtf^2}\right)+1\right).
    \label{eq:banerjee_penalty_fuzzy}
\end{equation}
 For the custom voxel weighting $ \alpha_s $ of the neighborhood term, we adapted \cref{eq:custom_neigh_weight} by using the standard deviation inside the neighborhood instead of the relative class occurrence:
\begin{equation}
 	\widetilde{\alpha}_{s}=
 	\begin{cases}
        \begin{aligned}
 		2\alpha, \quad &\text{if} \quad \sigma_{N_s}^{(1)}  < \sigma_{\text{thresh}} &\text{and} \quad \sigma_{N_s}^{(2)}  < \sigma_{\text{thresh}} \\
 		\alpha, \quad &\text{if} \quad \sigma_{N_s}^{(1)}  \geq \sigma_{\text{thresh}} &\text{and} \quad \sigma_{N_s}^{(2)}  < \sigma_{\text{thresh}} \\
 		\alpha, \quad &\text{if} \quad \sigma_{N_s}^{(1)}  < \sigma_{\text{thresh}} &\text{and} \quad \sigma_{N_s}^{(2)}  \geq \sigma_{\text{thresh}} \\
 		0, \quad &\text{otherwise} &\\
        \end{aligned}
 	\end{cases}
    \label{eq:neigh-weight_fuzzy}
 \end{equation}
 where $\sigma_{N_s}^{(k)}$ is the standard deviation in the $k$-th order neighborhood of voxel $s$ and $\sigma_{\text{thresh}}$ is a threshold for the standard deviation in the neighborhood of voxel $s$. With these adaptations, we can now reformulate \cref{eq:neigh_energy} to 
 \begin{equation}
     \Unf(\bm{X}=\bm{x})=\sum_{s\in S}\left(\alpha \widetilde{E}_{*}(x_s)\right)
    \label{eq:neigh-energy_fuzzy}
 \end{equation}
 where $\alpha$ can be either a fixed constant (normal neighborhood term) or a custom weight for each voxel (weighted neighborhood term), calculated using \cref{eq:neigh-weight_fuzzy}. $\widetilde{E}_{*}$ is calculated with \cref{eq:potts_penalty_fuzzy} or \cref{eq:banerjee_penalty_fuzzy}.

\subsection{HMRF-UNet}
\begin{figure}
    \centering
    \includegraphics[width=0.7\textwidth]{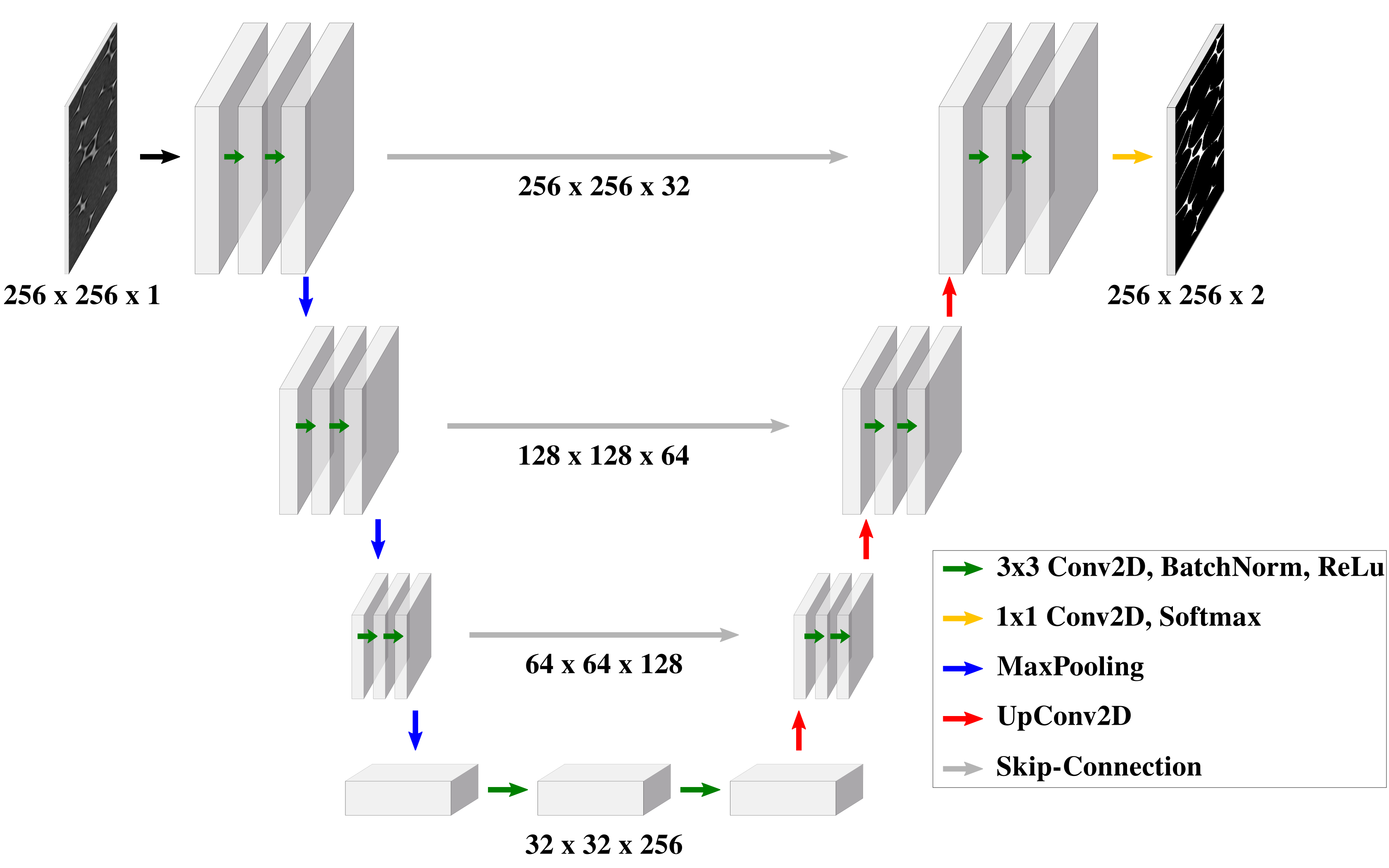}
    \caption{Overview of the Segmentation U-Net Architecture}
    \label{fig:U-Net}
\end{figure}

We used a vanilla U-Net for segmentation of the \textmu CT images. The U-Net consists of an encoder and a decoder path, where the corresponding encoder and decoder levels are connected using skip connections. In each level, several convolution blocks are used, each consisting of a \numproduct{3x3} Conv2D layer, a batch-normalization layer, and a ReLu layer. Between levels in the encoding path, a pooling layer with a pool size of 2 was used, while between levels in the decoding path, UpConv2D layers were used. The number of kernels per level doubled or halved with each level for the encoding and decoding path, accordingly. The final layer was a \numproduct{1x1} Conv2D layer with softmax activation and two output features. An exemplary representation of the architecture is shown in \Cref{fig:U-Net}.

The loss of the model consisted of two loss components, a distribution loss $\mathcal{L}_d$ and a neighborhood loss $\mathcal{L}_n$, which were weighted using the weights $\lambda_d$ and $\lambda_n$ accordingly:
\begin{equation}
    \mathcal{L} = \lambda_d \mathcal{L}_d + \lambda_n\mathcal{L}_n
    \label{eq:loss}
\end{equation}
The distribution loss $\mathcal{L}_d$ was calculated using \cref{eq:cond_llhood_fuzzy} and the neighborhood loss was calculated using \cref{eq:neigh-energy_fuzzy} based on a first-order neighborhood system with eight neighbors. For all the following studies, the distribution weight was set to $\lambda_d=1-\lambda_n$.

\subsection{Polyurethane foam datasets}
In this study, two different datasets were used. The first dataset (\textit{RealPUFoam}) contains real \textmu CT images of Polyurethane (PU) foam structures. The images have a size of \qtyproduct[product-units=single]{1300 x 951 x 960}{voxels} with a spatial resolution of \qtyproduct{2.75 x 2.75 x 2.75}{\um}. The second dataset (\textit{ArtPUFoam}) consisted of \qty{20000} artificially generated 2D \textmu CT images of PU foam structures with a size of \qtyproduct[product-units=single]{256 x 256}{pixels} and their ground-truth segmentations. The images of the artificial dataset were generated using a multistep workflow as described by \citet{griem_synthetic_2025}. In this workflow, a digital twin is created from an initial set of real \textmu CT images. The PACE3D simulation framework \cite{hotzer_parallel_2018} extracts structural features to generate similar binary foam structures. These binary foam structures are converted into grayscale images and paired into an artificial input image and a corresponding segmentation map to train a U-Net for image segmentation. The trained U-Net segments the original \textmu CT images, which are then used to train a generative adversarial network (GAN) for the generation of binary foams. Finally, a CycleGAN translates the GAN-generated binary images into grayscale images, guided by the structural characteristics of the real \textmu CT data.

\subsection{Data preprocessing}
\begin{figure}[bthp]
    \centering
    \includegraphics[width=\textwidth]{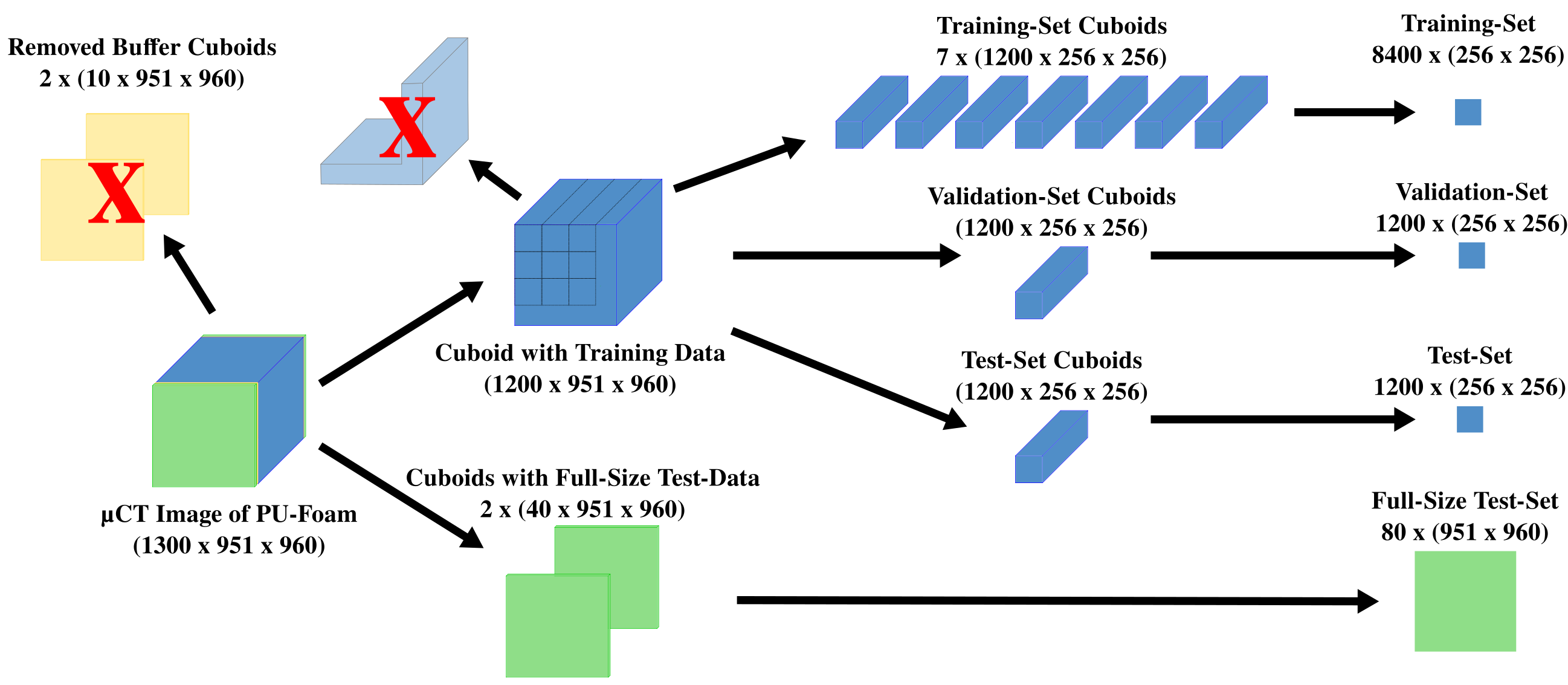}
    \caption{Visualization of the proposed cuboid-based dataset split. By extracting non-overlapping cuboids and splitting the data based on the cuboids, a leakage between training, validation and test set can be prevented.}
    \label{fig:GroupBasedSplit}
\end{figure}

Both datasets were preprocessed before being split into training, validation, and test set. For both datasets, the grayscale images were normalized to the range $[0, 1]$. Since the artificially generated dataset already contained large amounts of heterogeneous and independent 2D images, no further processing was necessary, and the images were split into training, validation, and testing sets with a split of $(70\%/15\%/15\%)$. Since the real PU foam images were 3D volumes, 2D slices had to be extracted. The adjacent slices of \textmu CT images share a lot of similarity, which could cause a leak between training, validation, and the test set. Therefore, we developed a cuboid-based dataset split as shown in \Cref{fig:GroupBasedSplit}. Similarly to the common group-wise split, the separation of cuboids prevents a leakage between the different datasets. We first removed 40 slices, from the front and back of the image, for an independent full-size test set. The following ten slices from front and back were discarded to prevent dataset leakage from the full-size test set to the training set. The cuboids were generated from the remaining volume with $1200$ slices using a non-overlapping moving window of size $1200\times256\times256$, resulting in nine separate cuboids. From each cuboid, slices of shape $256\times256$ were extracted. In addition, to increase the amount and heterogeneity of data, five different augmentations per slice were generated. A random contrast adjustment in the range $[0.8, 1.2]$ was combined with one of five transformations consisting of  $90^\circ$-rotation, $180^\circ$-rotation, $270^\circ$-rotation, $x$-axis-flip, and $y$-axis-flip. Overall this resulted in \qty{42000} training, \qty{6000} validation, and \qty{6000} test images.

\subsection{Network optimization and training}
The artificial dataset with pairs of images and ground-truth segmentations was used to find the optimal hyperparameters and to find the optimal loss setup. A first hyperparameter search for only network parameters was performed using Bayesian optimization with the Dice score (see \cref{sec:eval-metrics}) as optimization metric. Here, different values were tested for the size and starting number of kernels, levels, and convolutions per level, as well as type of pooling. The optimal resulting architecture was a U-Net with three levels of three convolutions each, max-pooling, a kernel size of $3\times3$, and 64 kernels in the top level. Using this fixed model setup, another Bayesian optimization was performed to find the optimal weight for the neighborhood loss term. The five best-performing values for the neighborhood loss weight were then used to train all models for further experiments. Based on empirical experiments, the learning rate was set to \qty{1e-5} and the batch size to 128 for all experiments. If not mentioned otherwise, all trainings were run for a total of 200 epochs.  All searches and trainings were performed using Tensorflow 2.11 on an Nvidia A100 with 80GB of GPU-RAM and 128 AMD EPYC 7513 32-core processors with an overall RAM of 2100GB.

\subsection{Evaluation metrics}
\label{sec:eval-metrics}
For the evaluation of the different models, which were trained using the \textit{ArtPUFoam} dataset, we used the commonly used Dice score (DSC) \cite{zijdenbos_morphometric_1994} as shown in \cref{eq:dsc} between the predicted segmentation $x$ and the ground-truth segmentation $\widehat{x}$. The Dice score values range from 0 (bad segmentation) to 1 (perfect segmentation). To ensure a fair comparison of models, a threshold sweep was run on the predicted confidence maps, to find the optimal threshold for creating the binary segmentation map, which was then used for the final calculation of the Dice score.   

\begin{equation}
    \mathrm{DSC}=\frac{2\big|x\cap \widehat{x}\big|}{\big|x\big|+\big|\widehat{x}\big|}
    \label{eq:dsc}
\end{equation}

\section{Results}
\label{sec:results}

\subsection{Numerical studies}
In a first study, the influence of the type and weight $\lambda_n$ of the neighborhood term on the segmentation performance was investigated using the \textit{ArtPUFoam} dataset. For this, we trained four different models with normal Potts neighborhood (HMRF-UNet\textsubscript{pot}), weighted Potts neighborhood (HMRF-UNet\textsubscript{wpot}), normal Banerjee neighborhood (HMRF-UNet\textsubscript{ban}), and weighted Banerjee neighborhood (HMRF-UNet\textsubscript{wban}) for the five different best-performing neighborhood loss weights from the hyperparameter search.
Another study investigated the influence of the parameter $\sigma_{\text{thresh}}$ in \cref{eq:neigh-weight_fuzzy}. For this HMRF-UNet\textsubscript{wban} models with $\lambda_n=0.31$ with $\sigma_{\text{thresh}}$ values of $0.05$, $0.10$, $0.15$, and $0.20$ were trained.
To test the usage of the HMRF-UNet for pre-training, a study was set up in which the performance of a supervised U-Net model trained with fractions of the \textit{ArtPUFoam} dataset with and without starting with pre-trained weights from an unsupervised HMRF-UNet was compared. The weights of the pre-trained model were extracted from an HMRF-UNet\textsubscript{pot} model with $\lambda_n=0.31$. For the supervised training of the U-Net we used the Dice loss suggested by  \citet{milletari_v-net_2016}.
In a final study, it was investigated how models trained on the \textit{RealPUFoam} dataset compare to models trained with the artifical \textit{ArtPUFoam} dataset. For this purpose, identical models with $\lambda_n=0.31$ with normal Potts neighborhood were trained both on the \textit{ArtPUFoam}, as well as on the \textit{RealPUFoam} dataset.

\subsection{Influence of neighborhood term}
\begin{figure}[bthp]
    \centering
    \includegraphics[width=\textwidth]{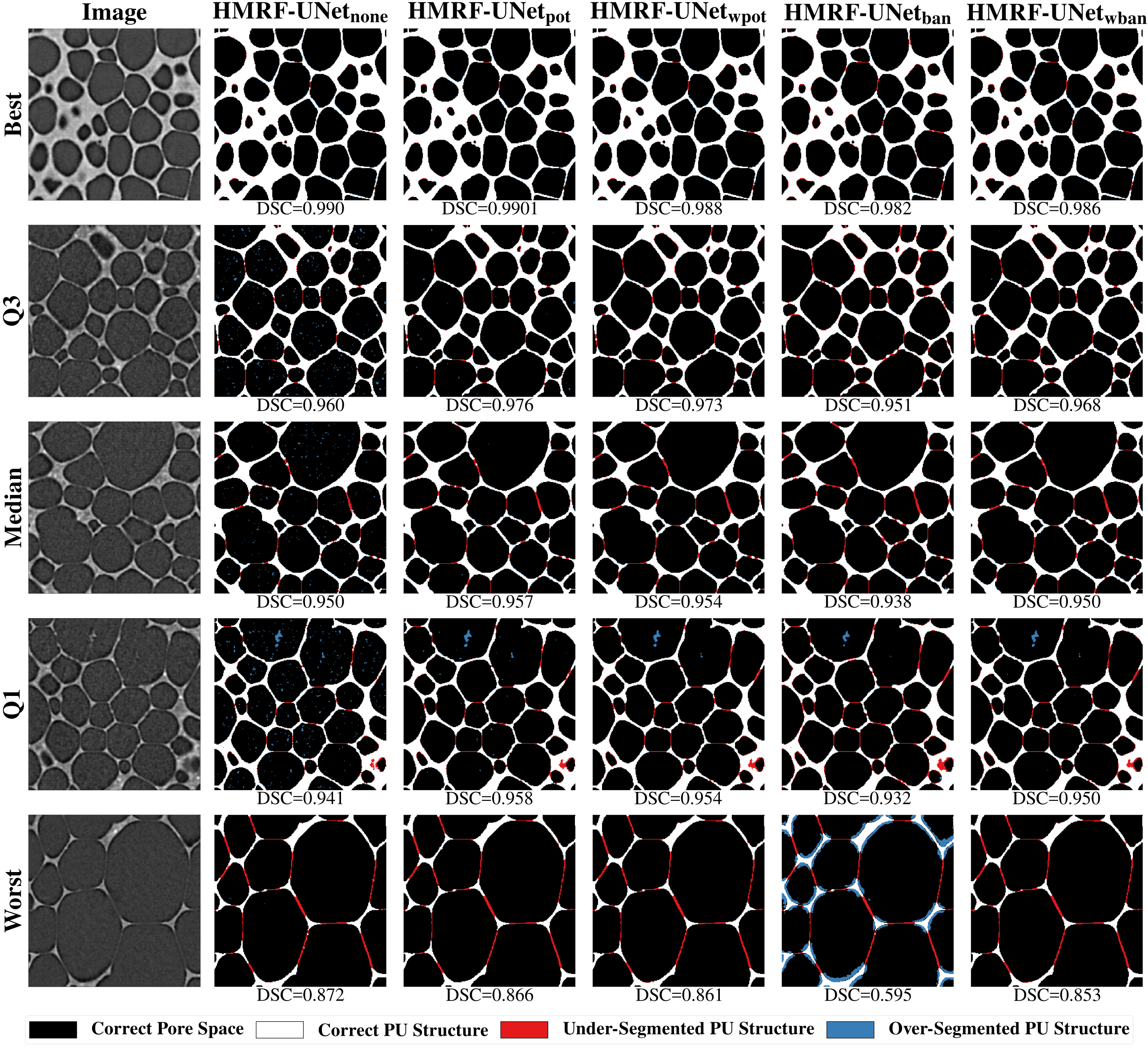}
    \caption{Comparison of segmentation results for different images from the testing set for five models trained with different neighborhood terms. The models trained with neighborhood term (columns 2-6) were trained with a neighborhood weight of $\lambda_n=0.31$. The model without neighborhood term (second column) over-segments the PU structure inside the pore space (blue areas). Training with the neighborhood term removes these over-segmentations. For the normal Banerjee neighborhood term an over-segmentation around the PU walls can be observed for the worst test image in the bottom row (blue areas). Otherwise most segmentation errors resulted from under-segmented PU structures (read areas).}
    \label{fig:NeighInfluence}
\end{figure}

The neighborhood loss can be defined by one of the four different neighborhood term choices (normal/weighted Potts and normal/weighted Banerjee). \Cref{tab:NeighInfluence} shows the average Dice scores for different combinations of neighborhood terms and weights. All models trained with the normal Banerjee term had considerably worse segmentations based on their average Dice score compared to all other models. The models with the Potts neighborhood had slightly higher Dice scores compared with the models with the weighted Banerjee neighborhood.  For custom weighted neighborhood terms, a decrease of the neighborhood-loss weight factor $\lambda_n$ led to an improvement of the segmentation, while the terms without custom weighting showed no influence of $\lambda_n$.
\Cref{fig:NeighInfluence} shows a visual comparison of the predicted segmentations of a model trained without a neighborhood term and the predicted segmentation of models trained with each of these neighborhood terms and a neighborhood weight of $\lambda_n=0.31$. In all cases, except for the worst image in the last row, the models with normal and weighted Potts neighborhood and the model with weighted Banerjee neighborhood generated better segmentations than the model without the neighborhood term. The model without the neighborhood term HMRF-UNet\textsubscript{noneigh} over-segmented the PU structure inside the pore space. We can also see that the model with normal Banerjee neighborhood always generated the worst segmentation result. For the bottom image, this was due to over-segmented PU structures around the PU walls (blue areas). The custom weighting led to improved segmentations for the Banerjee neighborhood formulation, but resulted in worse segmentations when used with the Potts neighborhood. For all models, the main segmentation errors were due to under-segmented PU structures (red areas). 

\begin{table}[bthp]
    \caption{Comparison of average Dice scores (Mean ± Std) for different neighborhood terms and weights.}
    \centering
    \begin{tabular}{l*{5}{S[table-format=1.3(3),detect-all]}}
    \toprule
     & \multicolumn{5}{c}{\bfseries Dice Score} \\
     \cmidrule{2-6}
     \bfseries NoNeighborhoodLoss & \multicolumn{5}{c}{$0.950\pm0.015$}\\
     \midrule
     \multicolumn{1}{r}{$\boldsymbol{\lambda_n}$}
     & \multicolumn{1}{c}{\bfseries 0.09} 
     & \multicolumn{1}{c}{\bfseries 0.18} 
     & \multicolumn{1}{c}{\bfseries 0.31} 
     & \multicolumn{1}{c}{\bfseries 0.35} 
     & \multicolumn{1}{c}{\bfseries 0.56} \\
    \midrule
    \bfseries NormalPottsLoss & 0.927(0.077) &  \bfseries 0.956(0.016) & \bfseries 0.957(0.017) & \bfseries 0.957(0.017) & \bfseries 0.951(0.019) \\
    \bfseries WeightedPottsLoss & \bfseries 0.956(0.015) & \bfseries 0.956(0.017) & 0.954(0.018) & 0.952(0.019) & 0.950(0.016) \\
    \bfseries NormalBanerjeeLoss & 0.952(0.015) & 0.914(0.091) & 0.882(0.103) & 0.882(0.103) & 0.771(0.088) \\
    \bfseries WeightedBanerjeeLoss & 0.955(0.016) & 0.953(0.018) & 0.947(0.019) & 0.946(0.019) & 0.943(0.013) \\    
    \bottomrule
    \end{tabular}
    \label{tab:NeighInfluence}
\end{table}

\subsection{Threshold for custom neighborhood-weighting}
\begin{figure}[bthp]
    \centering
    \includegraphics[width=\textwidth]{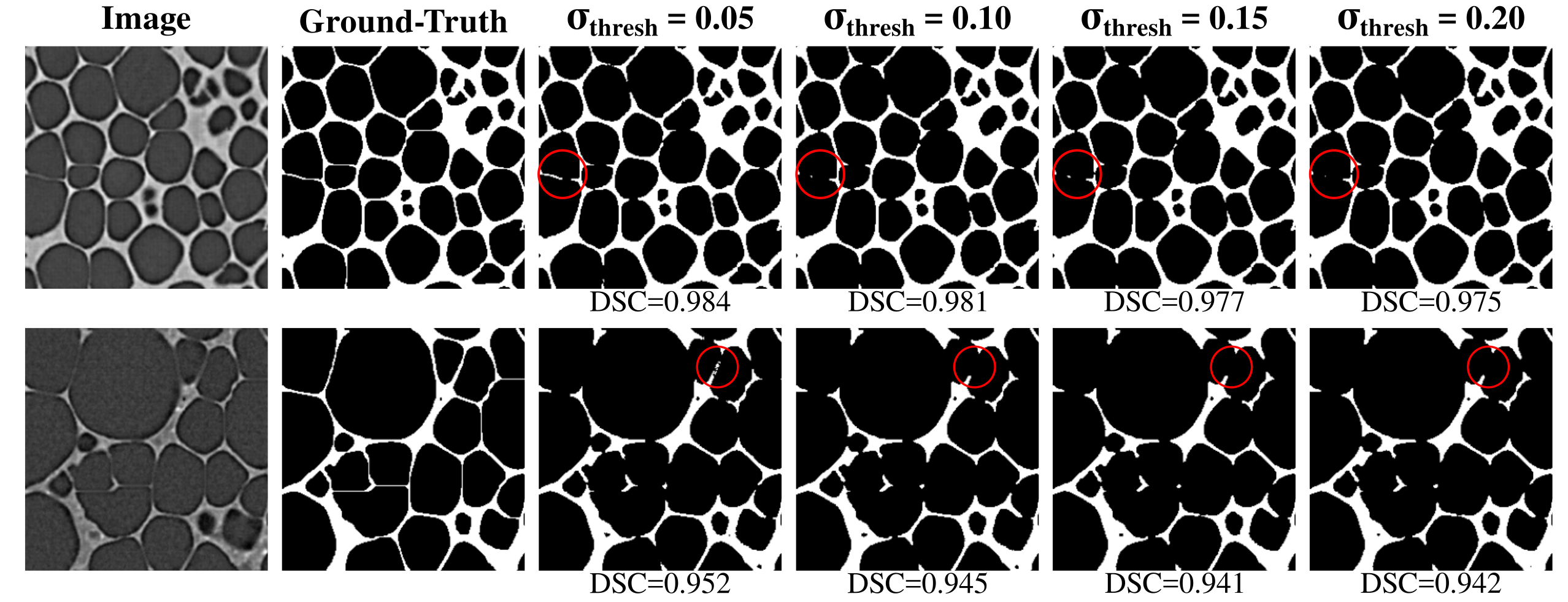}
    \caption{Comparison of segmentation results of models trained with weighted Banerjee neighborhood with different thresholds $\sigma_{\text{thresh}}$ for two exemplary images. In the regions marked with red circles, differences of segmentation accuracy become apparent, showing improved segmentation of thin structures for smaller thresholds.}
    \label{fig:SigmaThresh_Influence}
\end{figure}

An analysis revealed the influence of the threshold $\sigma_{\text{thresh}}$ used for the calculation of the custom neighborhood weight. \Cref{fig:SigmaThresh_Influence} shows the influence $\sigma_{\text{thresh}}$ for two exemplary images of the test set. For both images, a reduction of the threshold led to improved segmentation results. Especially, thin PU struts were partially detected for the smaller thresholds of $0.05$, which were not detected using higher thresholds of $0.20$, as can be seen in the areas marked with red circles in \Cref{fig:SigmaThresh_Influence}. 

\subsection{HMRF-UNet for pre-training}
\begin{figure}[bthp]
    \centering
    \includegraphics[width=\textwidth]{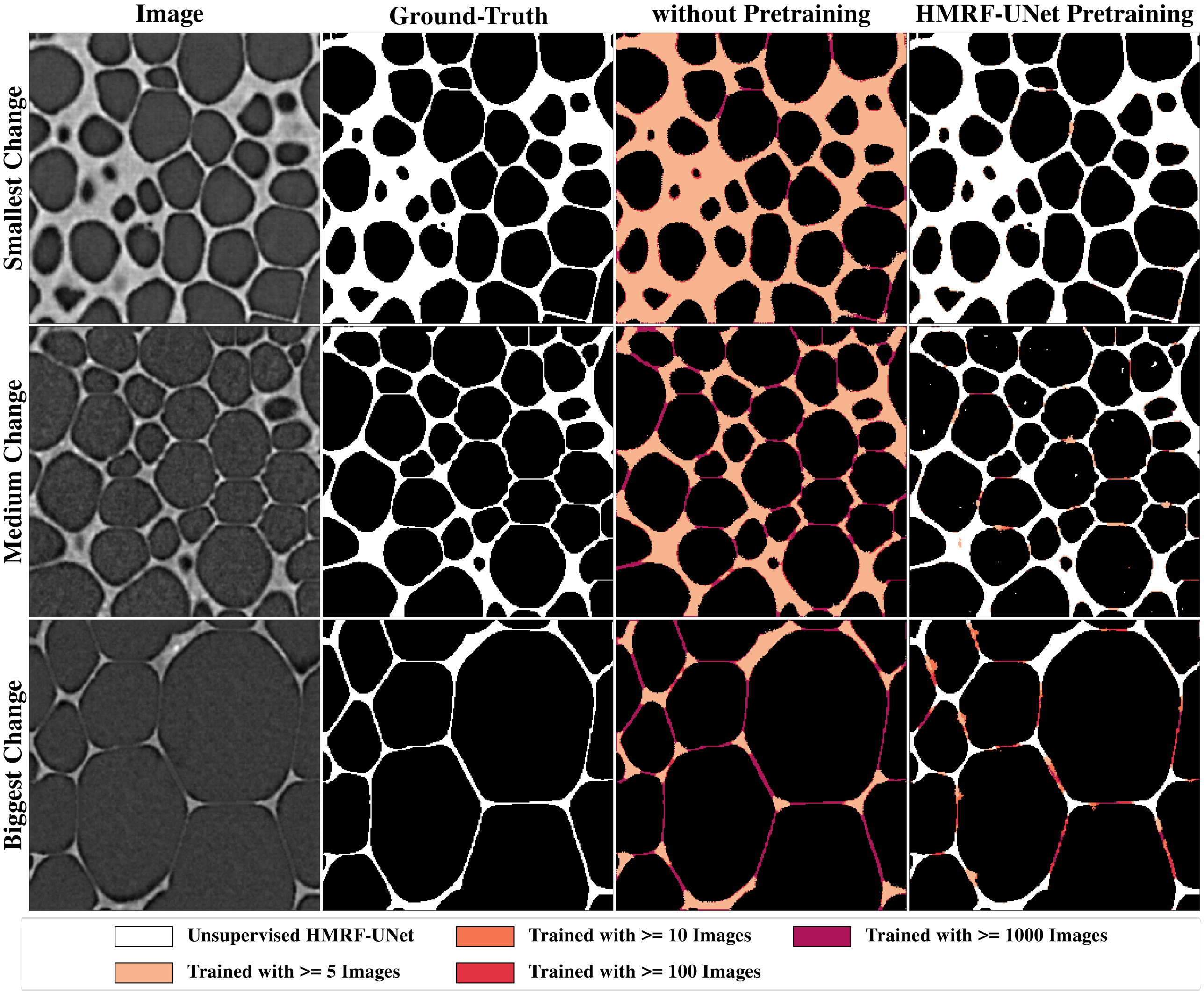}
    \caption{Effect of using HMRF-UNet as a pretrained model for finetuning with different amount of ground-truth pairs using supervised loss. The grayscale image (first column), the ground-truth segmentation (second column), the segmentation of the supervised model without pretraining (third column) and the segmentation of the supervised model with HMRF-UNet pretraining are shown for examples of the test set with the smallest change (first row), median change (second row) and biggest change (third row). The colour coding of the segmentations in the third and fourth column show the improvement of using more images for fine-tuning.}    \label{fig:PretrainingComparison}
\end{figure}

\begin{figure}[bthp]
    \centering
    \includegraphics[width=0.6\textwidth]{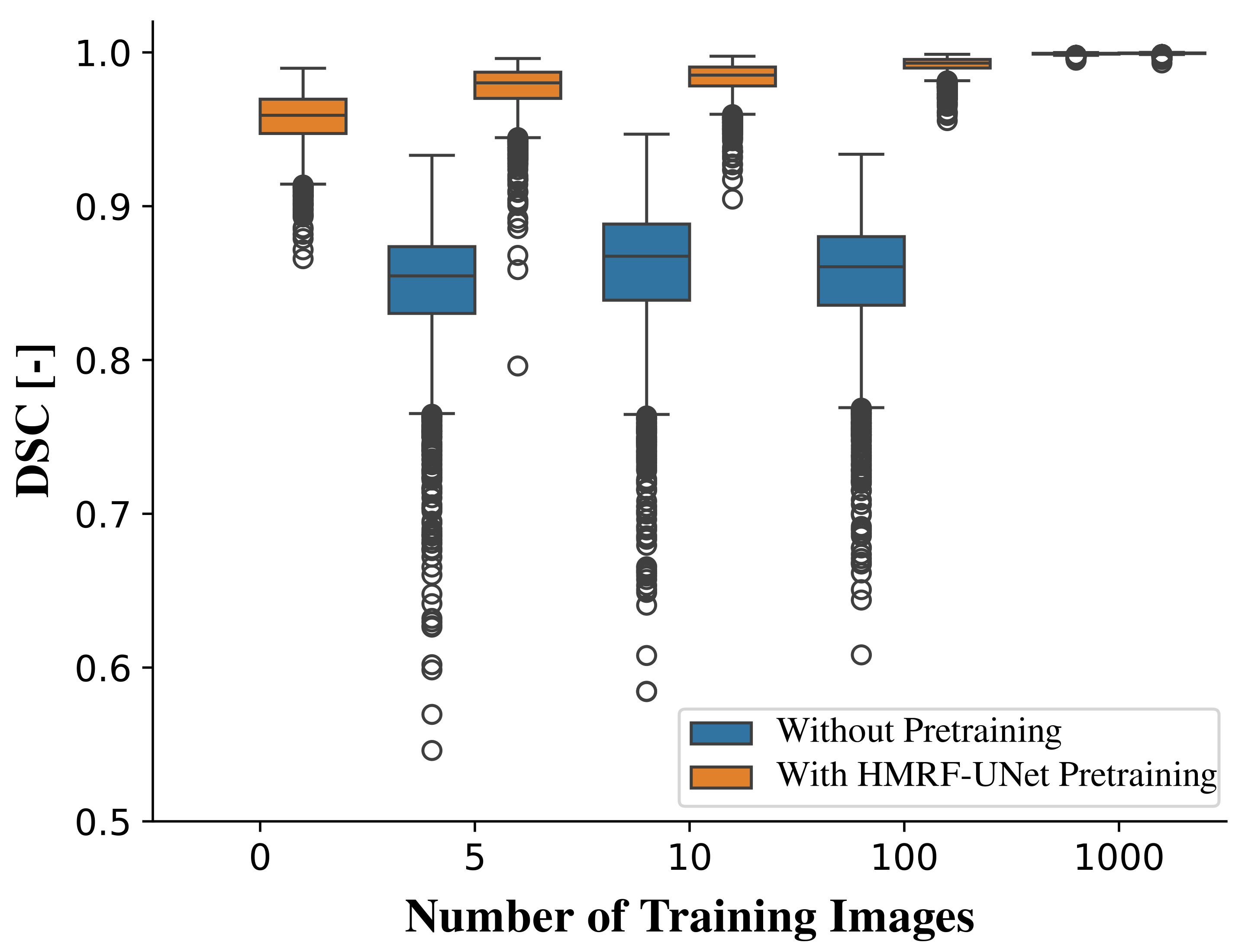}
    \caption{Boxplots comparing the Dice scores on the test set for the supervised models without HMRF-UNet pretraining (blue) and the supervised models with HMRF-UNet pretraining (orange). For all amount of training images, the pre-training significantly (p<0.001) improved the Dice scores (DSC) of the segmentation.}
    \label{fig:DSC_Pretraining}
\end{figure}

\Cref{tab:HMRF-Pretraining} compares models, which were trained using a supervised Dice loss with and without using the weights of a model trained with unsupervised HMRF-loss as a starting configuration for different amounts of training data. It can be observed that for the models without any pre-training, only the models trained with at least 1000 images achieve high Dice scores. The models with pre-training already achieved high Dice scores for the smallest training set of only five images. The supervised training improved the Dice scores of the unsupervised HMRF-UNet from $0.957$ to $0.977$, $0.983$ and $0.992$, when fine-tuning on five, ten or 100 images, respectively. The effect of pre-training can also be observed in \Cref{fig:PretrainingComparison}, which compares the segmentation results of the images of the test set with the smallest (first row), median (second row) and biggest change (third row) for supervised models with and without pre-training using the HMRF-UNet. The overlay plots in the third and fourth columns show how fine-tuning with more images improved the segmentation. It can be observed that the unsupervised HMRF-UNet already generated good segmentations (white structures). Most of the missing fine PU walls could be recovered after fine-tuning with ten images. When fine-tuning with 100 supervised examples, we achieved near-perfect segmentation results. For the supervised model without pre-training (third column), a training with 1000 images was necessary to successfully segment the fine PU walls. The boxplots of the Dice scores of both model variants in \Cref{fig:DSC_Pretraining} confirm these observations. For all amounts of fine-tuning/training data, the model with HMRF-UNet pre-training outperformed the model without pre-training. The significance of these differences was confirmed by a Wilcoxon signed-rank test, which resulted in p-values of p<0.001 for all five cases.

\begin{table}[tbh]
    \caption{Analysis of the influence of using a pretrained HMRF-UNet as a starting point for supervised learning of segmentation tasks. Comparison of average Dice scores (Mean $\pm$ Std) for models trained with and without pre-training for different supervised training set sizes.}
    \centering
    \begin{tabular}{l*{5}{c}}
    \toprule
     & \multicolumn{5}{c}{\bfseries Dice Score} \\
     \cmidrule{2-6}
     \bfseries Training Image Amount& 
     \bfseries 5 &
     \bfseries 10 &
     \bfseries 100 &
     \bfseries 1000 &
     \bfseries 14000 (all)\\
    \midrule
    \bfseries Unsupervised  &
    \multirow[c]{2}{*}{-} &
    \multirow[c]{2}{*}{-} &
    \multirow[c]{2}{*}{-} &
    \multirow[c]{2}{*}{-} &
    \multirow{2}{*}{$0.957\pm0.017$}\\
    \bfseries HMRF-UNet\\
    \midrule
    \bfseries Supervised U-Net &  \multirow[c]{2}{*}{$0.848\pm0.040$} & \multirow[c]{2}{*}{$0.859\pm0.043$} &  \multirow[c]{2}{*}{$0.854\pm0.038$} &  \multirow[c]{2}{*}{$0.999\pm0.001$}  &  \multirow[c]{2}{*}{$1.000\pm0.000$}  \\
    \bfseries without Pre-Training \\
    \cmidrule{1-1}
    \bfseries Supervised U-Net & \multirow[c]{2}{*}{$0.977\pm0.015$} & \multirow[c]{2}{*}{$0.983\pm0.010$} &   \multirow[c]{2}{*}{$0.992\pm0.005$}  &  \multirow[c]{2}{*}{$0.999\pm0.000$}  &  \multirow[c]{2}{*}{-}  \\ 
    \bfseries with HMRF Pre-Training \\
    \bottomrule
    \end{tabular}
    \label{tab:HMRF-Pretraining}
\end{table}

\subsection{Real \textmu CT validation}
\begin{figure}[bthp]
    \centering
    \includegraphics[width=\textwidth]{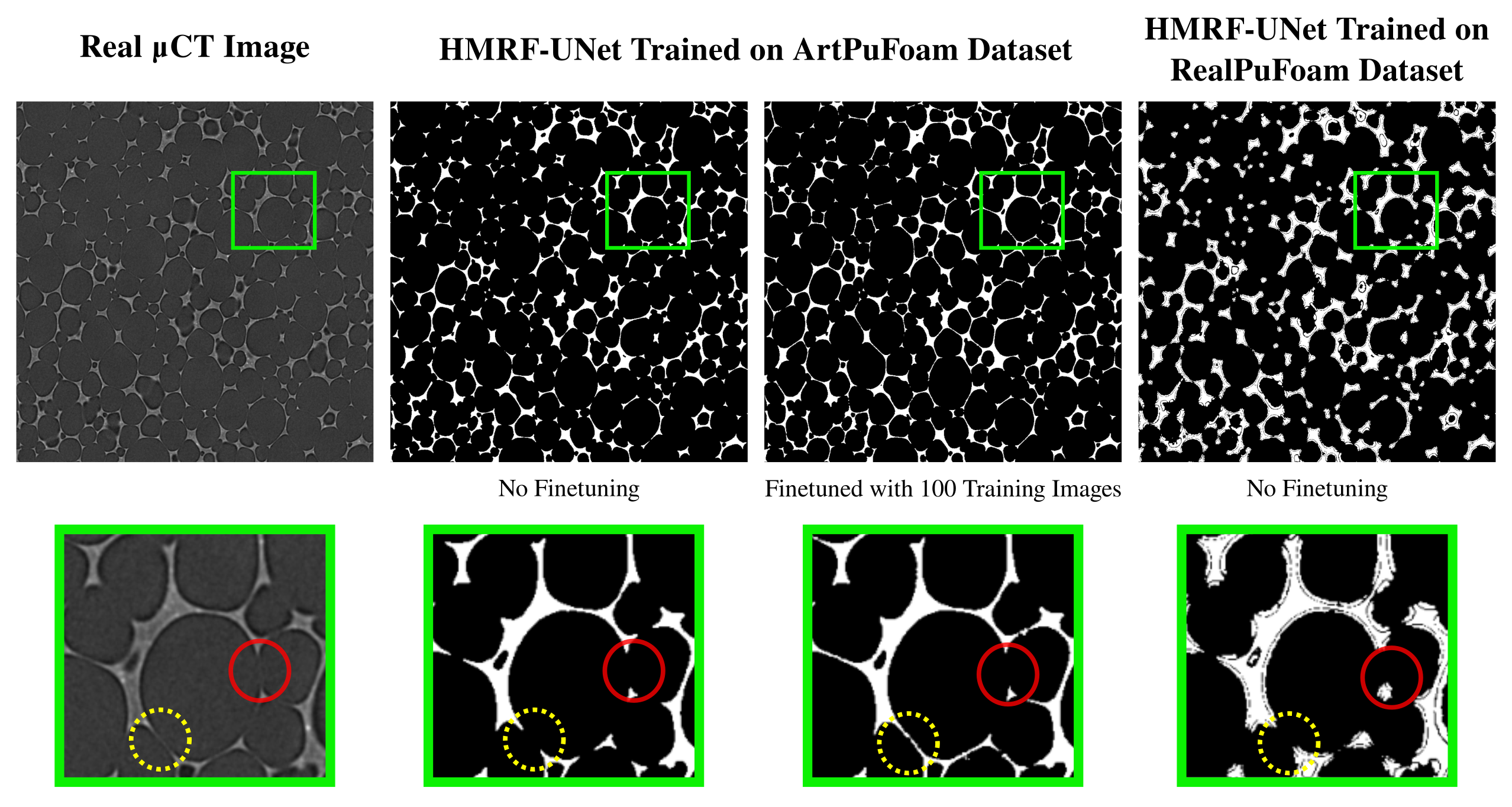}
    \caption{Comparison of predicted segmentations for an exemplary image from the \textit{RealPUFoam} dataset for different models trained with regular Potts neighborhood and $\lambda_n=0.31$. Top row: Full Segmentation, Bottom Row: Enlarged regions of interest. The HMRF-UNet trained on the \textit{RealPUFoam} generates worse segmentations compared to the two models trained on the \textit{ArtPUFoam} dataset. Red circles indicate an area where all models missed fine PU walls, yellow dotted circles mark areas where finetuning improves the segmentation of thin walls.}
    \label{fig:RealDataComparison}
\end{figure}

To validate the segmentation performance of the HMRF-UNet on the real \textmu CT dataset, three different models all with normal Potts neighborhood and $\lambda_n=0.31$ were trained. The first model was trained on the \textit{ArtPUFoam} dataset, the second model was based on the first model but finetuned with 100 ground-truth images. The third model was trained on the \textit{RealPUFoam} dataset without any finetuning afterwards. The predicted segmentations for an exemplary full-size test image are shown in \Cref{fig:RealDataComparison}. It can be observed that some thin PU walls were never segmented (red circles), while other thin PU walls were at least segmented by the fine-tuned model (yellow dotted circle). It also becomes apparent that the model trained on the real dataset generated over-segmented PU structures and generated the visually worst segmentation results.

\section{Discussion}
\label{sec:discussion}

\subsection{Choice of neighborhood term}
We have shown that the choice of neighborhood loss definition has an important influence on the segmentation quality of the model. In general, the experiments suggest that both Potts neighborhood types outperform the newer Banerjee neighborhood types. A reason for this might be that for a binary segmentation the regular Banerjee neighborhood does not bring any benefit over the Potts neighborhood, since calculating the difference of the means of a class is not too different from calculating the distance between the confidence map predictions. Another possible reason might be that the mean and standard deviations in the original implementation of the Banerjee neighborhood are based on non-fuzzy, fixed class assignments and are therefore just the properties of one class. In our case, the means and standard deviations are, however, calculated using a weighting with the confidence value of a voxel. 

It could also be observed that custom neighborhood weighting is especially important for the Banerjee neighborhood loss. Without custom weighting, the segmentation results were
remarkably worse compared to the segmentation results from the models with custom weighting. The exemplary segmentations in \Cref{fig:NeighInfluence} suggest that the low Dice scores may be contributed to over-segmentations for input images with lower contrast. The contrast in the input image appeared to influence not only the segmentation quality of the HMRF-UNet\textsubscript{ban}, but also all the different models, since for all four models the Dice score was remarkably lower for the image with low contrast in the last row. To test this hypothesis, we calculated the standard deviation inside the input images as a measure of the contrast of the image. The average standard deviations of the 50 worst and 50 best predictions for the HMRF-UNet\textsubscript{ban} model were calculated and resulted in an average standard deviation of $0.187\pm0.020$ for the best predictions and $0.097\pm0.011$ for the worst predictions. This observation supports our hypothesis that a lower image contrast leads to worse segmentations.     
The choice of the right threshold value in the weighting term can also help to improve the segmentation of small structures in the image, as shown in \Cref{fig:SigmaThresh_Influence}.

\subsection{Weaknesses of the HMRF loss}
When investigating the segmentations obtained with the HMRF-UNet, we could observe that the model often struggles with the thin walls of the PU foam structure. One reason for this are the low intensities of some of the small walls. This could lead to possible assignments to the darker background/air class, since the voxel intensity fits better to the air class distribution. In supervised approaches, the model receives feedback from the ground-truth labels, which enables these models to even learn the segmentation of structures which are hardly visible in the image itself. This is a challenge, which most unsupervised methods share. Another reason for missing some of the small walls might be the neighborhood loss. Since these thin walls are often only one voxel thin, there are only two neighbors belonging to the same PU class. Since most of the neighbors belong to the air class, the neighborhood loss punishes the model if it classifies the voxel as PU structure. 

The redundant class problem described by \citet{nan_unsupervised_2022} could also influence the quality of the segmentation. We set the number of classes of the HMRF model to the expected number of unique classes, which is two in our case. However, some of the voxels in our image might jump between these two classes. Therefore, using more than two classes could be beneficial, allowing these uncertain voxels to be assigned to their own class. Also, artifacts like dark shadows, or bright points influence the properties of the distribution of their class, which could lead to wrong distribution properties. If these artifacts get assigned to their own, additional class, this could allow to ignore these artifacts. 

\subsection{Effectiveness of unsupervised pre-training}
The positive effect of unsupervised pre-training due to a regularization effect was shown before \cite{erhan_why_2010,caron_unsupervised_2019,kalapos_self-supervised_2023}. Our results suggest that a pre-training using the unsupervised HMRF loss can reduce the required amount of ground truth data for training segmentation models. We could show that a fine-tuning of the HMRF-UNet with only five to ten images significantly improved the segmentation quality, and fine-tuning with 100 ground-truth images led to near-perfect segmentation performance on the \textit{ArtPUFoam} dataset. This is in good agreement with the results of \citeauthor{kalapos_self-supervised_2023}, who showed that they could achieve near-perfect segmentations by fine-tuning a pre-trained model with only 100 ground truth images \cite{kalapos_self-supervised_2023}. This demonstrates the potential of unsupervised HMRF loss for any segmentation task, suggesting it as a viable alternative or supplementary pre-training method to common approaches such as clustering, contrastive learning, and generative models.

\subsection{Training with artificial vs. real dataset}
When comparing the segmentation results for the real \textmu CT images in \Cref{fig:RealDataComparison}, it could be observed that the HMRF-UNet trained on the \textit{RealPUFoam} dataset surprisingly performed worse than the model trained on the \textit{ArtPUFoam} dataset. The oversegmentations could be explained by a dark border between the pore space and the PU walls. This dark border has extremely low intensities, which are far away from the mean intensities of both classes. Since we predefined the number of classes to two, the dark borders had to be merged with one of the other two classes. The mentioned dark borders were not present in the artificial dataset, and therefore, the predictions of the models trained with this dataset did not generate over-segmentations. The overall worse segmentation quality for the real dataset could be attributed to two reasons. Again, the dark borders around PU walls likely also influence the prediction and prevent the correct segmentation of thin PU walls. The second reason could be partial volume artifacts, which might be present in the real \textmu CT and lead to mixed signals of PU and air. This effect might be especially pronounced in the thin PU walls, which could lead to worse segmentation accuracy in these areas.

\subsection{Limitations}
Our study has several limitations. For the training of the model, only pixel intensities were used. We did not use any textural features. In addition, we only tested our proposed HMRF-UNet on real and artificial \textmu CT data of PU foams. Since PU foam structures consist of only two classes, the segmentation task was easier, compared to images with more than two classes. Furthermore, only  ground-truth data for the artificial dataset was available. Therefore, the performance of our model could only be quantitatively evaluated for this artificial dataset and not for the dataset with real \textmu CT images, where only a qualitative evaluation was possible. In general, we want to emphasize that this study was a proof-of-concept study to show the potential of the HMRF-UNet. More experiments should be carried out to explore additional influencing factors and possible optimizations in the loss function of other model components.

\section{Conclusion}
\label{sec:conclusion}
In this work, the application of HMRF theory in constructing an HMRF loss for unsupervised segmentation using a U-Net was demonstrated. By combining the advantages of Hidden Markov Random Fields and CNN segmentation, we trained an unsupervised model that predicts segmentation maps in around 200 ms. The investigation of the loss components showed the importance of the neighborhood loss and the choice of suitable neighborhood loss configurations. It was also demonstrated how pre-training with HMRF loss can significantly reduce the required amount of ground-truth data for training well-performing segmentation networks. To our knowledge, this was the first time an HMRF loss was included in a CNN for unsupervised end-to-end segmentation. 
The findings suggest that the HMRF-UNet is another interesting and useful unsupervised segmentation approach that complements other unsupervised approaches, such as contrastive learning and SOMs.    

Other authors have shown that using a combination of supervised and unsupervised loss terms can also improve the performance of the model \cite{rasmus_semi-supervised_2015}. Therefore, a follow-up study could investigate the combination of the unsupervised HMRF loss with a supervised Dice loss for training of a U-Net for segmentation.
We also plan to further investigate the combination of other unsupervised learning approaches, such as contrastive learning, with the HMRF loss to explore whether this can further improve segmentation performance in a complete unsupervised setting. Further studies could also investigate whether the usage of more than two classes improves the segmentation accuracy. We aim to validate the HMRF-UNet on other, more diverse datasets from material science and medicine. With only minor changes, the HMRF-UNet could also be trained on 3D image data. Further studies should evaluate whether a 3D HMRF-UNet, which incorporates 3D neighborhood systems, improves the segmentation of the real \textmu CT images, especially for thin PU walls.

\section*{Acknowledgments}
The study was funded by the Deutsche Forschungsgemeinschaft (DFG, German Research Foundation) - \#NE822/31. Contributions have also been funded by the Helmholtz association within the joint lab MDMC and the program MSE, no. 43.31.01 which is gratefully acknowledged. 

\section*{Data Availability}
The used image datasets and the scripts for training the HMRF-UNet are publicly available at\\https://doi.org/10.5281/zenodo.17590658.

\bibliographystyle{unsrtnat}  
\bibliography{references}

\end{document}